\documentclass[journal]{IEEEtran}

\ifCLASSINFOpdf
\else
   \usepackage[dvips]{graphicx}
\fi
\usepackage{url}

\usepackage{graphicx}
\usepackage{amssymb}
\usepackage{float} 
\usepackage{subfigure}
\usepackage{booktabs}

\begin{document}

\title{Ghost-free High Dynamic Range Imaging via Hybrid CNN-Transformer and Structure Tensor}

\author{Yu Yuan$^\dagger$, Jiaqi Wu$^\dagger$, Zhongliang Jing,~\IEEEmembership{Senior Member,~IEEE}, Henry Leung,~\IEEEmembership{Fellow,~IEEE}, Han Pan
	\thanks{	Manuscript received September 6, 2022. This work was supported by the National Natural Science Foundation of China (Grant Nos. 61175028, 61603249, 61673262)	
		(Corresponding authors: Zhongliang Jing, and $\dagger$ indicates equal contribution.) }
	\thanks{Yu Yuan, Zhongliang Jing, and Han Pan are with the School of Aeronautics and Astronautics, Shanghai Jiao Tong University, Shanghai 200240, China (e-mail:cascyy@sjtu.edu.cn; zljing@sjtu.edu.cn; hanpan@sjtu.edu.cn).}
	\thanks{Jiaqi Wu  is with the School of Information and Communication Engineering, University of Electronic Science and Technology of China (UESTC), Chengdu 610097, China (e-mail: dl2wjq@gmail.com). }
	\thanks{Henry Leung is with the	Department of Electrical and Computer Engineering, University of Calgary, Calgary, AB T2N 1N4, Canada (e-mail: leungh@ucalgary.ca).}
}

\markboth{IEEE SIGNAL PROCESSING LETTERS, Vol. 66, No. 66, Octoember 2022}
{Shell \MakeLowercase{\textit{et al.}}: Bare Demo of IEEEtran.cls for IEEE Journals}
\maketitle

\begin{abstract}
Eliminating ghosting artifacts due to moving objects is a challenging problem in high dynamic range (HDR) imaging. In this letter, we present a hybrid model consisting of a convolutional encoder and a Transformer decoder to generate ghost-free HDR images. In the encoder, a context aggregation network and non-local attention block are adopted to optimize multi-scale features and capture both global and local dependencies of multiple low dynamic range (LDR) images. The decoder based on Swin Transformer is utilized to improve the reconstruction capability of the proposed model. Motivated by the phenomenal difference between the presence and absence of artifacts under the field of structure tensor (ST), we integrate the ST information of LDR images as auxiliary inputs of the network and use ST loss to further constrain artifacts. Different from previous approaches, our network is capable of processing an arbitrary number of input LDR images. Qualitative and quantitative experiments demonstrate the effectiveness of the proposed method by comparing it with existing state-of-the-art HDR deghosting models. Codes are available at https://github.com/pandayuanyu/HSTHdr.

\end{abstract}

\begin{IEEEkeywords}
High dynamic range deghosting, hybrid, CNN-Transformer, structure tensor.
\end{IEEEkeywords}

\IEEEpeerreviewmaketitle

\section{Introduction}

\IEEEPARstart{T}{he} dynamic range of natural scenes can be very wide, spanning several orders of magnitude from dazzling sunlight to faint starlight. Most camera systems can only capture a very narrow dynamic range, which is often reflected in the fact that we get a picture with over-exposed or under-exposed regions. To address aforesaid limitation, many high dynamic range (HDR) imaging techniques have been developed. One way is to use a  specialized camera to capture HDR images \cite{hardware1,hardware2}.
However, the cost of using these devices is high for general consumers.
Another common strategy is to use an exposure bracketing with different exposure values (EV) to compose the final HDR image \cite{sig97,mertens,mti}. These approaches work well for static scenes, however, when the scenes have moving objects or the camera is slightly displaced, these methods suffer from severe ghosting artifacts [see Fig. \ref{structuretensor}(b)]. 

\begin{figure}[htb]
	\centering
	\includegraphics[width=3.5in]{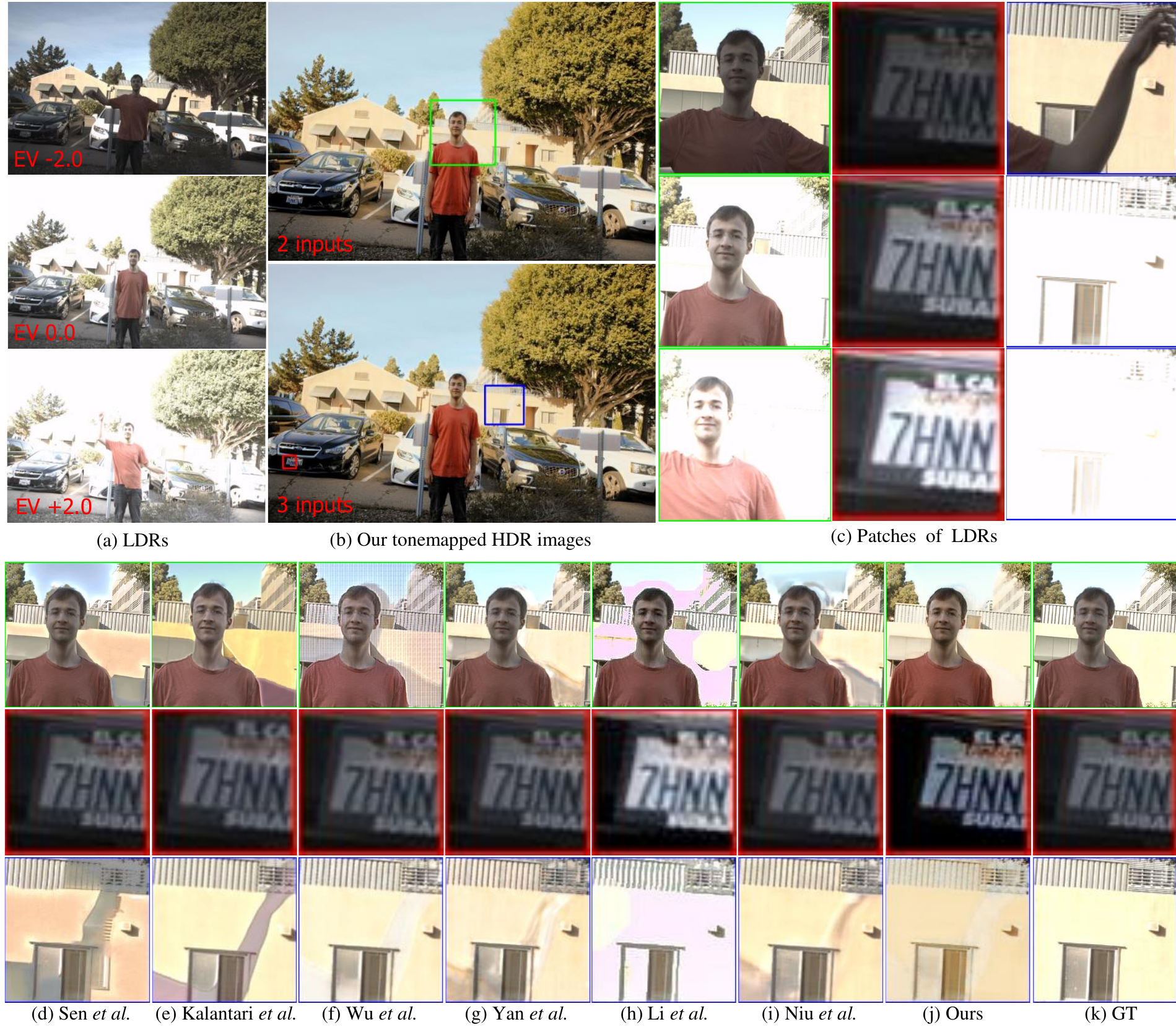}
	\caption{The input LDR images with large motions are shown in (a). The first two images (EV -2.0 and EV 0.0) are fed into our network to obtain the HDR image shown in the top of (b), this image continues to be fused with the remaining LDR image (EV +2.0) to obtain the result in the bottom of (b).  Other methods require three LDR images as inputs, and to compare with our dual-input network, we duplicate the second image to replace the original third LDR image. }
	\label{xiaoguo}
\end{figure}

Many conventional methods have been proposed to generate ghost-free HDR images. They can be roughly divided into alignment before merging methods and  motion rejection methods. The former approach aligns the LDR images before merging them. \cite{bogoni,kang,zimmer} used optical flow to estimate the motions. These optical flow-based methods rely heavily on the accuracy of the estimation. Many studies use the PatchMatch algorithm \cite{patchmatch} to pre-align LDR images. Sen  \textit{et al.} \cite{robust} proposed a patch-based energy minimization method and used a joint optimization strategy to integrate the alignment and HDR reconstruction processes.  Hu \textit{et al.} \cite{hu} optimized the aligned LDR images based on intensity mapping function and gradient consistencies on the transformed domain. However, these methods are not very effective for large displacements and have a high computational cost. For the motion rejection methods, they first align the input LDR images at the overall level, and then reject unaligned pixels. Gallo \textit{et al.} \cite{gallo} proposed to predict the color in various exposures and compare it with the original values to detect motions. Khan \textit{et al.}  \cite{khan} proposed to assign fusion weights by calculating the probability of a given pixel belonging to the background. Heo \textit{et al.} \cite{heo} proposed to detect unaligned regions by calculating the joint probability density between LDR images. Lee \textit{et al.} \cite{rank1} employed rank minimization to generate ghost-free images. A common drawback of these methods is that they ignore useful information about the abandoned pixels.

Because of its strong representational capacity, a myriad of deep learning-based HDR deghosting methods have emerged in recent years. Kalantari \textit{et al.} \cite{Kalantari} introduced the first multi-frame HDR imaging method for dynamic scenes based on deep learning. This method feeds the input LDR images into the fusion network after aligning them by optical flow. Wu \textit{et al.} \cite{wu} used homography to globally align the LDR inputs and then fed them into a UNet-based network for local alignment and fusion. Yan \textit{et al.} \cite{ahdr} used a attention mechanism to suppress the non-aligned regions between the input LDR images. Niu \textit{et al.} \cite{hdrgan} proposed the first GAN-based method for HDR deghosting.  

Existing HDR deghosting architectures are mostly tailored for fixed-number (e.g., three) input LDR images and use one of them as the reference image. This setting greatly limits the practicality of these methods, since we can only obtain 2 LDR images with different exposures or more than three in many scenarios. Meanwhile, ghosting artifacts are more pronounced in the gradient field \cite{gradident1}, but are strongly affected by noise. It is shown that structure tensor  \cite{ST, STTV}  is able to reflect the textures of images while remaining largely free from noise interference. These aforementioned findings inspire us to develop a new network capable of processing an arbitrary number of input images, and introduce the structural tensor information of images into our network to better constrain ghosting artifacts.

The main contributions of our works are summarized as follows.
\begin{enumerate}
	\item{We develop a hybrid CNN-Transformer network to process dual-input LDR images for HDR deghosting. A contextual aggregation network and non-local channel attention block are utilized in the encoder to extract multi-scale features and establish local and global dependencies within and between the source images, respectively. A sequential fusion strategy is explored to accommodate an arbitrary number of inputs.}
	\item{We aggregate the structural tensor information of LDR images into the network and introduce the structural tensor loss to further constrain the ghosting artifacts.}
	\item{We conduct comparative experiments on publicly available datasets. Qualitative and quantitative evaluations validate the superiority of our method over conventional methods.}
\end{enumerate}

\section{The proposed method}

\begin{figure}[htb]
	\centering
	\includegraphics[width=3.5in]{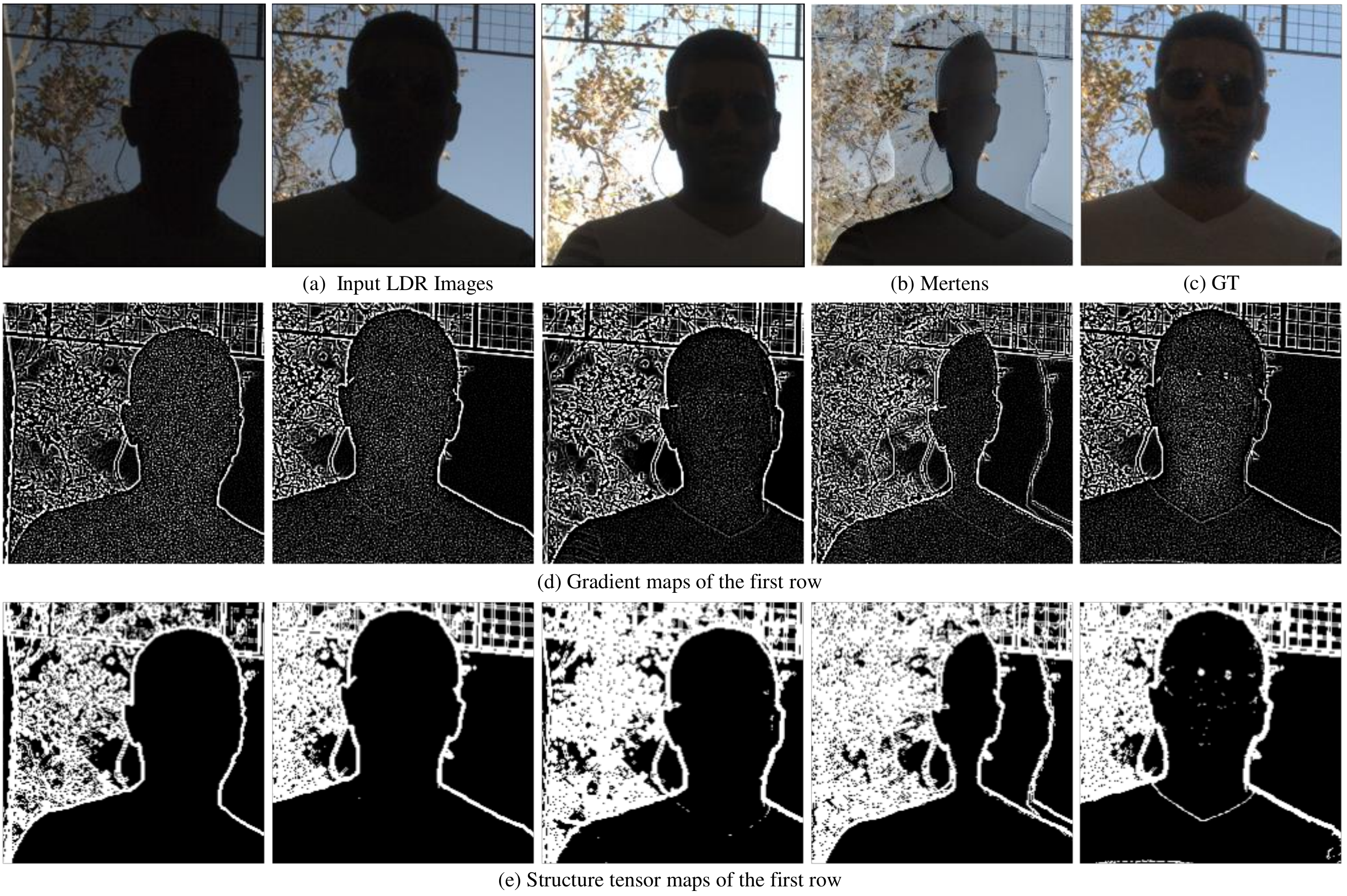}
	\caption{The difference between gradient and structure tensor. Figure (a) shows the three LDR images of the input. Figure (b) gives the fusion result obtained by Mertens \textit{et al.}, which shows the typical ghosting patterns. }
	\label{structuretensor}
\end{figure}

\begin{figure*}[htb]
	\centering
	\includegraphics[width=6.9in]{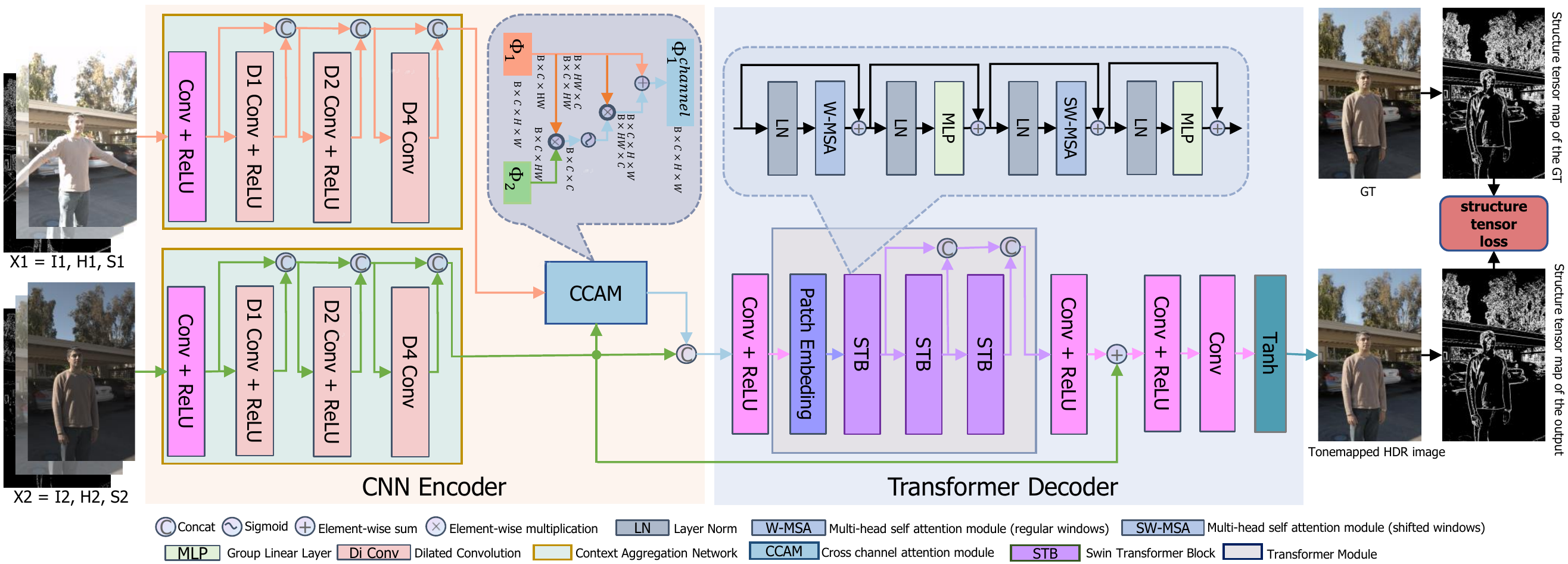}
	\caption{Overview of the proposed HDR deghosting framework.}
	\label{framework}
\end{figure*}

This section describes the proposed network based on hybrid CNN-Transformer and  structure tensor for HDR deghosting.
Our goal is to convert an LDR image sequence $\left\{I_{1}, I_{2}, \ldots, I_{k}\right\}$ into a high-quality HDR image,  $H$. In our network, we input only two images (e.g. $I_{1}, I_{2}$) at a time and use the resulting image as the input for an arbitrary number of input images in subsequent tasks. 
In \cite{Kalantari}, images in the LDR domain are used to detect noise and saturated regions, while images in the HDR domain facilitate detection of motions.
HDR image $H_{k}$ can be obtained from LDR images by the following formula:
\begin{equation}
 	H_{k}=\frac{I_{k}^{\gamma}}{t_{k}}, \quad k=1,2
\end{equation}
where $t_{k}$ denotes the exposure time of the $k$-th image $I_{k}$. $\gamma$ is a
gamma correction parameter, which is set to 2.2. Following \cite{Kalantari}, we concatenate $I_{k}$ and $H_{k}$ along the channel dimension to obtain a 6-channel input $X_{k}=\left\{I_{k}, H_{k}, i=1,2\right\}$. 
Therefore, the process of obtaining HDR images can be expressed as:
\begin{equation}
	H=\mathbf{M}\left(X_{1}, X_{2}; \theta\right)
\end{equation}
where $\mathbf{M}(\cdot)$ represents the network, and $\theta$ denotes the network parameters.

 Since artifacts are more significant in the gradient domain  \cite{gradident2, gradident3} [see the fourth and fifth columns of Fig. \ref{structuretensor}], and using gradient information can effectively mitigate ghosting artifacts. However, we find that the gradient of images is sensitive to noise, which will introduce many interference [see Fig. \ref{structuretensor}(d)]. As shown in Fig. \ref{structuretensor}(e), the detection of flat areas in the structure tensor \cite{ST,STTV,yu} allows us to obtain structural information about images while reducing the effects of noise. Comparing the gradient maps with the structure tensor maps, we find that the results obtained by the structure tensor have clearer textures, which shows the potential of the structure tensor for deghosting. In summary, we  concatenate the inputs of the network with structural tensor information $S_{k}$, and get the progressive 7-channel input, $X_{k}=\left\{I_{k}, H_{k}, S_{k}, k=1,2\right\}$.

\subsection{Proposed Framework for HDR Deghosting}
As illustrated in Fig. \ref{framework}, our network adopts a hybrid CNN-Transformer structure, which is mainly composed of a CNN encoder and a Transformer decoder. We set the second image $X_2$ as the reference. The CNN encoder is responsible for extracting multi-scale features of the inputs and learning motions between $X_2$ and $X_1$. The Transformer decoder is used to reconstruct ghost-free HDR images.

We use the context aggregation network (CAN) \cite{CAN} with different dilation rates (here we set 1, 2, 4) to obtain multi-scale 64-channel features $\Phi_{1}$ and $ \Phi_{2}$ without losing spatial resolution. Considering the degradation of the spatial properties of the deep features extracted by CAN, we only utilize a channel attention block. By extending the non-local attention mechanism \cite{Nonlocal} to multiple heterogeneous inputs, we propose a cross channel attention module (CCAM) to learn the difference between $X_1$ and the reference $X_2$ by establishing associations between $\Phi_{1}$ and $\Phi_{2}$ in latent space. The $\Phi_{2}$ and  $\Phi_{1}^{channel}$ obtained by CCAM are concatenated and then fed into the Transformer decoder. 

Zhao \textit{et al.} \cite{hybrid} demonstrates that the hybrid Transformer-CNN architecture has a better performance in terms of both model capacity and computational complexity. Therefore, we use the Swin transformer block proposed in \cite{SwinT} as part of the decoder. Specifically, three successive Swin Transformer blocks are utilized and concatenated to obtain features of different depths. It is worth noting that  a skip connection between the encoder and decoder is added to prevent network degradation.

\begin{figure}[htb]
	\centering
	\includegraphics[width=3.4in]{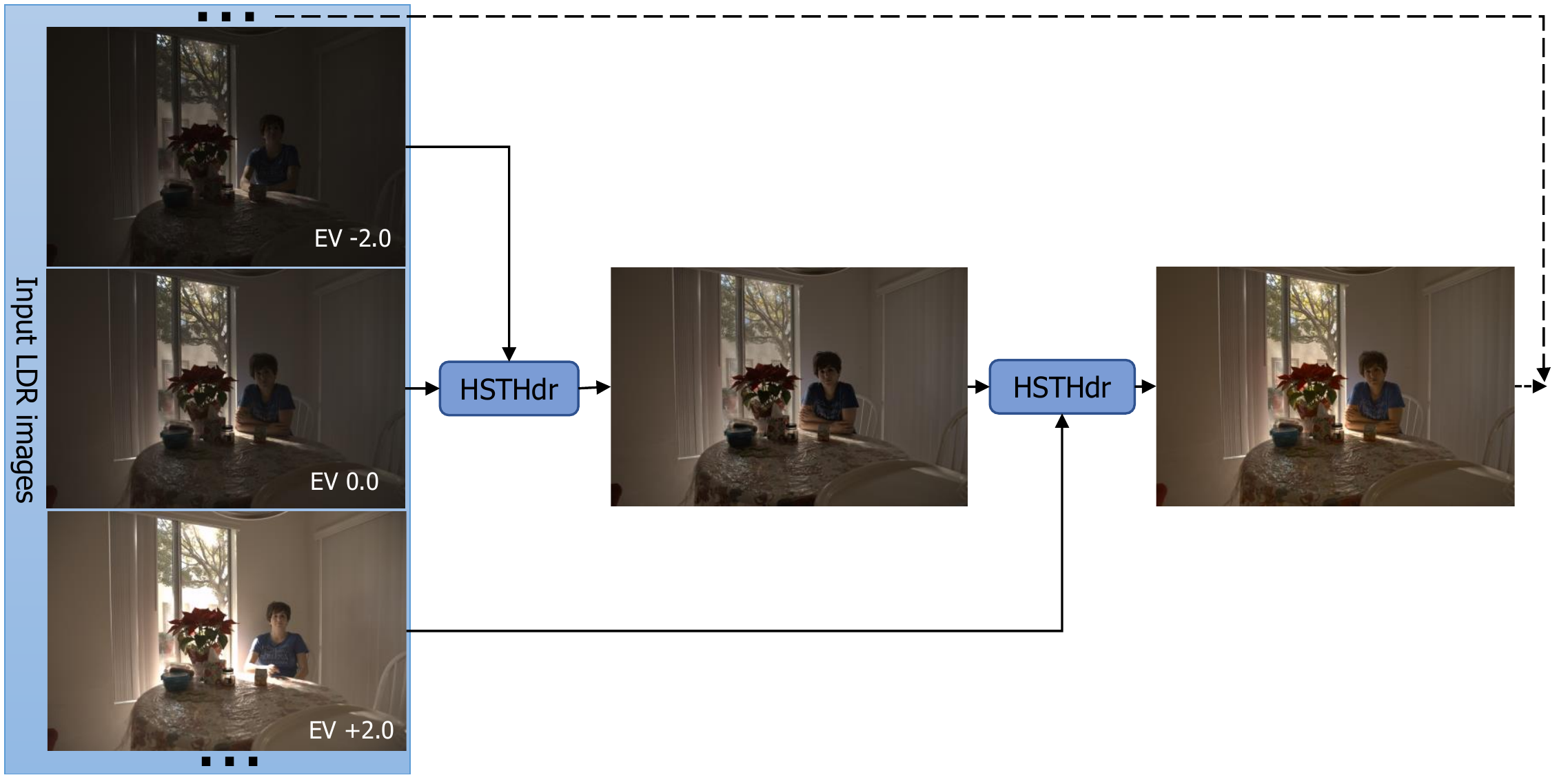}
	\caption{Proposed sequential fusion strategy in our approach.}
	\label{input}
\end{figure}

\subsection{Loss Function}
Due to the low dynamic range of our regular consumer-grade displays, HDR format images need to be tonemapped to have a better visual experience on these monitors. According to \cite{Kalantari}, $\mu$-law is used for tonemapping:
\begin{equation}
	T(H)=\frac{\log (1+\mu H)}{\log (1+\mu)}
\end{equation}
where $\mu$ is set to 5000. $T(H)$ is the tonemapped result of HDR image $H$.  

Our loss function is defined as:

\begin{equation}
	\mathcal{L}=\|T(H)-T(G T)\|_{2}^{2}+\lambda\left\|T\left(H_{S T}\right)-T\left(GT_{S T}\right)\right\|_{1}
\end{equation}
where $GT$ denotes the ground truth, $H_{ST}$ represents the structure tensor map of $H$, and $GT_{ST}$ is the structure tensor map of the ground truth. $\|\cdot\|_{2}^{2}$ denotes MSE loss, and $\|\cdot\|_{1}$ refers to L1 loss. $\lambda$ indicates the weight of structure tensor loss.

\subsection{Dealing with Arbitrary Number of Inputs}

As shown in Fig. \ref{input}, for an LDR sequence with an arbitrary number of inputs, we initially take the image with middle exposure value (EV 0.0) in the sequence as the reference and fed it into our proposed dual-input network together with the LDR image of lower exposure value. The intermediate image obtained by our network is used as the new reference and is fused with another image of higher exposure value. We observe that the strategy of fusing the reference image with the darker image first and then with the brighter image has a  better performance. It is probably because the network tends to give more weight to pixels with larger luminance values.

\section{Experiments}

\label{eva}

In this section, we perform extensive experiments to validate the effectiveness of our proposed model. We use the dataset in \cite{Kalantari}, specifically by splitting a set of three LDR images into two sets that share the same reference image. We generate 150 samples of training data and 30 samples for validation. Our network is trained with an AdamW optimizer. The initial learning rate is set as 1$e$-4 and decreases to 1$e$-8 with the cosine annealing strategy. we select $ \lambda$ = 1$e$-2  as the weight of structure tensor loss. Random seeds are fixed during all training sessions.  We train for 5000 epochs with batch size 4 and set the input image resolution as 256×256 pixels. All experiments are performed on one NVIDIA Geforce RTX 3090 GPU and Intel Core i9-10900k CPU @ 3.70GHz. Our network is programmed on PyTorch.

Four metrics are utilized to quantify the fusion results, including peak signal-to-noise ratio (PSNR) \cite{PSNR} after tonemapping using $\mu$-law (PSNR-$\mu$) and in the linear domain (PSNR-$l$). We also use a quality metric developed specifically for the multi-exposure fusion of dynamic scenes MEF-SSIMd \cite{mefssimd}. In addition, to evaluate the subjective quality of HDR images, we use neural image assessment (NIMA)\cite{NIMA}.
\begin{table}[H]
	\caption{\textbf{Quantitative results of ablation studies on 9 validation image pairs. Values in bold indicate the best results.}}
	\label{ablation}
	\centering
	\setlength{\tabcolsep}{1.1mm}{
		\begin{tabular}{l|cccc}
			\toprule%第一道横线
			                       &   PSNR-$l$ (dB)  &   PSNR-$\mu$  (dB)  & MEF-SSIMd     &  NIMA    \\
			\midrule%第二道横线 
			w/o STB        & 39.7832   &     36.2759      &   0.9182   &    4.3445    \\
			w/o ST   &   36.0576     &  32.8702       &   0.8901     &  4.0742         \\
			w/o CCAM               &  37.3947    &	33.7846        &  0.8997     &  4.1970       \\
			\midrule
			Ours                  &\textbf{39.9299} &\textbf{37.3522}&\textbf{0.9196}& \textbf{4.3983} \\
			\bottomrule%第四道横线
	\end{tabular}}
\end{table}

\begin{figure}[htb]
	\centering
	\includegraphics[width=3.5in]{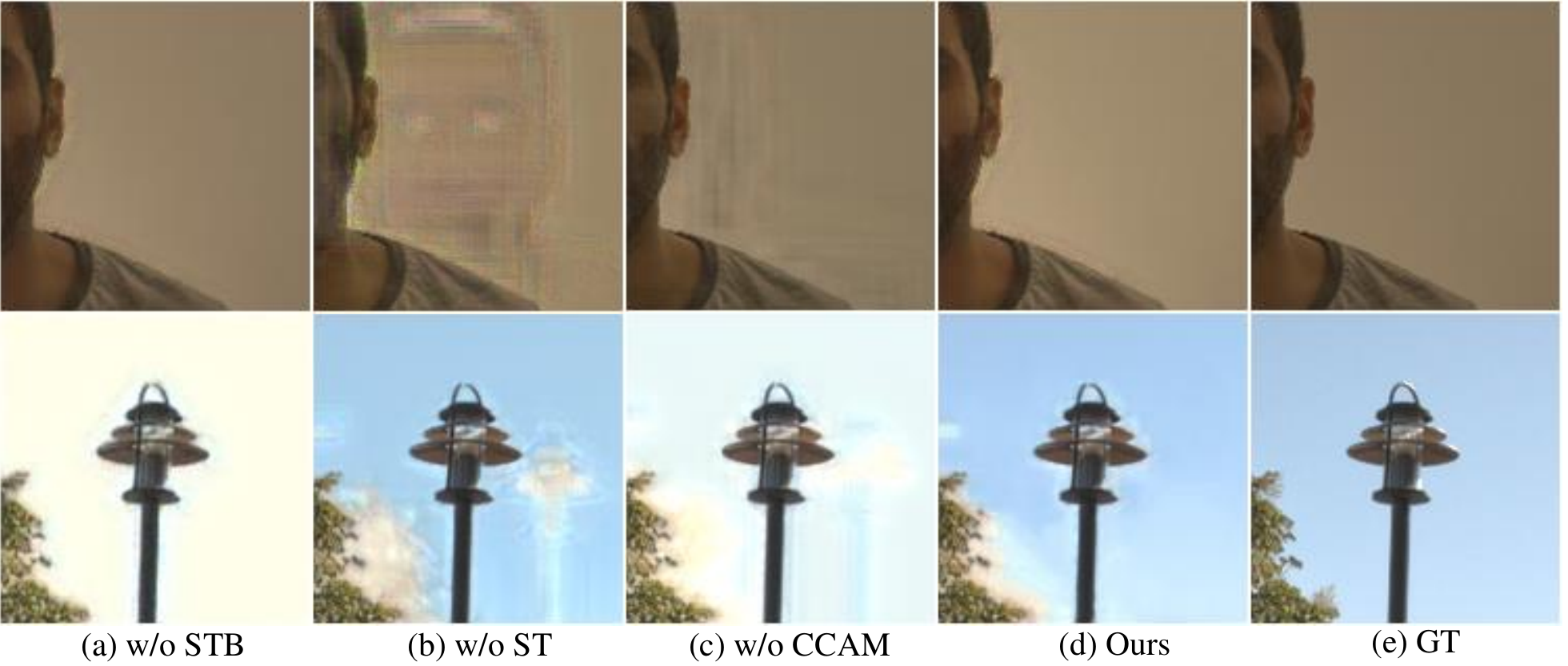}
	\caption{Subjective comparisons of ablation studies. (a) w/o Swin Transformer Blocks; (b) w/o Structure Tensor; (c) w/o CCAM; (d) ours; (e) GT.}
	\label{xiaorong}
\end{figure}

\subsection{Ablation Studies}

{\textit{Swin Transformer Blocks:}} To verify the contribution of the Transformer module, the Swin Transformer blocks are replaced by dilated residual dense block (DRDB) proposed by \cite{ahdr}. As shown in Table \ref{ablation}, the hybrid CNN-Transformer scheme outperforms the pure CNN scheme by about 0.15db and 1.1db in PSNR-$\mu$ and PSNR-$l$, respectively, and also overtakes in MEF-SSIMd and NIMA.  Fig. \ref{xiaorong} shows that our method is more realistic in terms of color and details.

{\textit{Structure Tensor:}} We remove the structural tensor information from the network inputs and loss functions to explore the necessity of using it. Table \ref{ablation} demonstrates that the use of structure tensor effectively improves the quality of fused images. Fig. \ref{xiaorong} illustrates that without the constraint of the structure tensor, the obtained HDR images suffer from significant ghosting artifacts.

{\textit{Cross Channel Attention Module:}} We abandon the proposed cross-channel attention module (CCAM) as a way to explore its impact on the elimination of ghost artifacts. Table \ref{ablation} demonstrates that the presence of CCAM can effectively improve the quality of the output HDR images. As shown in Fig. \ref{xiaorong}  , the proposed CCAM can also suppress ghosting artifacts to a certain extent.

\subsection{Comparison With Other HDR Deghosting Algorithms}

The comparative study is performed with six state-of-the-art HDR deghosting methods, one patch-based method (Sen \textit{et al.} \cite{robust}) , one flow-based CNN method (Kalantari \textit{et al.} \cite{Kalantari}) , two CNN based method( Wu \textit{et al.} \cite{wu}, Yan \textit{et al.} \cite{ahdr}), one multi-exposure fusion method ( Li \textit{et al.} \cite{fmmef}), and a GAN-based method (Niu \textit{et al.} \cite{hdrgan}). 10 validation image samples in \cite{Kalantari} are used for comparisons.

\begin{table}[H]
	\caption{\textbf{Average metric values with different HDR deghosting methods. Values in bold indicate the best results. $-$ means that Li \textit{et al.}  can not produce HDR images in $.hdr$ format}}
	\label{dabijiao}
	\centering
	\setlength{\tabcolsep}{1.6mm}{
		\begin{tabular}{l|cccc}
			\toprule%第一道横线
		             Method          	& PSNR-$l$ & PSNR-$\mu$  &   MEF-SSIMd  &  NIMA \\
			\midrule%第二道横线 
			 Sen \textit{et al.}        &  41.8625 &  37.1662 &  0.9006   &  4.5661 \\
	    	Kalantari \textit{et al.}   &  33.8925 & 28.2505  &  0.7393   &  4.6170 \\
     	 Wu	\textit{et al.}             &  34.1839 & 29.2874  &  0.8213   &  4.5393 \\
	  	Yan	\textit{et al.}             &  40.1744 & 36.9848  &  0.9212   &  4.4591 \\
			Li	\textit{et al.}         &      -   &	-     &  0.8221   &  4.1515 \\
			Niu \textit{et al.}         &  36.7448 & 34.5755  &  0.8629   &  4.5583 \\
			\midrule
			Ours (2 inputs)       & 40.0943 &36.9922 &\textbf{0.9245}& 4.6163    \\
			Ours  (3 inputs)    & \textbf{43.5397} &\textbf{38.9873} &	0.9233 & \textbf{4.6257}    \\
			\bottomrule%第四道横线
	\end{tabular}}
\end{table}

Table \ref{dabijiao} shows the results when three LDR images are input to the methods for comparison, as well as the results for our method when only the first two and all three images of the sequence are input. Our method achieves the best performance on PSNR-$l$ and PSNR-$\mu$ when three LDR images are input. Notably, our method leads in the NIMA, which means that our method can satisfy human visual sense. Interestingly, our method scores higher on the MEF-SSIMd with only 2 inputs than with 3 inputs.  From Fig. 1, our method has an advantage at two inputs, and also achieves the least ghosting artifacts and the clearest textures among the results obtained by the sequence fusion strategy of three-input.

\section{Conclusion}

In this letter, we propose a novel ghost-free high dynamic range imaging architecture based on a hybrid CNN-Transformer en/decoder network. The structural tensor information introduced in the network can mitigate ghost artifacts more effectively. The dual input model and the sequential fusion strategy make our method independent of the number of inputs. Evaluation experiments verify the effectiveness of our proposed scheme compared with other state-of-the-art HDR deghosting models.

\end{document}